# Variational Algorithms for Marginal MAP

**Qiang Liu**
Department of Computer Science
University of California, Irvine
Irvine, CA, 92697
qliu1@ics.uci.edu

**Alexander Ihler**
Department of Computer Science
University of California, Irvine
Irvine, CA, 92697
ihler@ics.uci.edu

## Abstract

Marginal MAP problems are notoriously difficult tasks for graphical models. We derive a general variational framework for solving marginal MAP problems, in which we apply analogues of the Bethe, tree-reweighted, and mean field approximations. We then derive a "mixed" message passing algorithm and a convergent alternative using CCCP to solve the BP-type approximations. Theoretically, we give conditions under which the decoded solution is a global or local optimum, and obtain novel upper bounds on solutions. Experimentally we demonstrate that our algorithms outperform related approaches. We also show that EM and variational EM comprise a special case of our framework.

## 1 INTRODUCTION

Graphical models provide a powerful framework for reasoning about structured functions defined over many variables. The term *inference* refers generically to answering probabilistic queries, such as computing probabilities or finding optima. Although NP-hard in the worst case, recent algorithms, including variational methods such as mean field and the algorithms collectively called belief propagation can approximate or solve these problems in many practical circumstances.

Three classic inference tasks include *max-inference* problems, also called maximum *a posteriori* (MAP) or most probable explanation (MPE) problems, which look for the most likely configuration. A second type are *sum-inference* problems, which calculate marginal probabilities or the distribution's normalization constant (the probability of evidence in a Bayesian network). Finally, *marginal MAP*, *mixed*, or *max-sum-inference* tasks seek a partial configuration of variables that maximizes those variables' marginal probability.

These tasks are listed in order of increasing difficulty: max-inference problems can be shown to be NP-complete, while sum-inference is #P-complete, and mixed-inference is $\text{NP}^{\text{PP}}$-complete (see e.g. Park and Darwiche, 2004). Practically speaking, max-inference tasks have a number of efficient algorithms such as loopy max-product BP, tree-reweighted BP, MPLP, and dual decomposition methods (see e.g., Koller and Friedman, 2009; Sontag et al., 2011). Sum-inference problems are similarly well-studied, and a set of algorithms parallel to those for max-inference also exist.

Perhaps surprisingly, mixed max-sum inference is much harder than either max- or sum- inference problems alone. A classic example illustrating this is Fig. 1, where marginal MAP in a simple tree structure is still NP-hard (Koller and Friedman, 2009). The difficulty is caused in part because the max and sum operators do not commute, so their order is not exchangeable. For this reason, research on marginal MAP is still relatively unexplored, with a few exceptions, e.g., Doucet et al. (2002), Park and Darwiche (2004), Huang et al. (2006), and Jiang and Daumé III (2010).

In this paper, we extend the concepts of variational approaches to the marginal MAP problem, enabling a host of techniques to be applied. These lead to new, powerful algorithms for estimating and bounding the marginal MAP solutions, for which some global or local optimality conditions can be characterized. We relate our algorithms to existing, similar approaches and validate our methods in experimental comparisons.

## 2 BACKGROUND

Graphical models capture the factorization structure of a distribution over a collection of variables. Let

$$p(x;\theta) = \exp\bigl(\theta(x) - \Phi(\theta)\bigr), \quad \theta(x) = \sum_{\alpha \in \mathcal{I}} \theta_\alpha(x_\alpha) \quad (1)$$

$$\text{and} \quad \theta = \{\theta_\alpha\}, \quad \Phi(\theta) = \log \sum_x \exp \theta(x)$$

where $\alpha$ indexes subsets of variables, and $\Phi(\theta)$ is the normalizing constant, called the *log-partition function*. We associate $p(x)$ with a graph $G = (V, E)$, where each variable $x_i$, $i = 1 \ldots n$ is associated with a node $i \in V$ and $(ij) \in E$ if $\{i, j\} \subseteq \alpha$ for some $\alpha$. The set $\mathcal{I}$ is then a set of cliques (fully connected subgraphs) of $G$. In this work we focus on pairwise models, in which the index set $\mathcal{I}$ is the union of nodes and edges, $\mathcal{I} = V \cup E$.

## 2.1 SUM-INFERENCE METHODS

Sum-inference is the task of marginalizing (summing out) variables in the model. Without loss of generality, it can be treated as the problem of calculating the log partition function $\Phi(\theta)$. Unfortunately, straightforward calculation requires summing over an exponential number of terms.

Variational methods are a class of approximation algorithms that transform inference into a continuous optimization problem, which is then typically solved approximately. To start, we define the *marginal polytope* $\mathbb{M}$, the set of marginal probabilities $\tau = \{\tau_\alpha(x_\alpha) | \alpha \in \mathcal{I}\}$ that correspond to a valid joint distribution, i.e.,

$$\mathbb{M} = \{\tau : \exists \text{ distribution } q(x), \text{ s.t. } \tau_\alpha(x_\alpha) = \sum_{x \setminus x_\alpha} q(x)\}$$

For any $\tau \in \mathbb{M}$, there may be many such $q$, but there is a unique distribution of form (1), denoted $q_\tau$, with maximum entropy $H(x; \tau) = -\sum_x q_\tau(x) \log q_\tau(x)$. We write $H(x; q_\tau)$ as simply $H(x; \tau)$ for convenience. A key result to many variational methods is the convex dual form of the log-partition function,

$$\Phi(\theta) = \max_{\tau \in \mathbb{M}} \langle \theta, \tau \rangle + H(x; \tau), \qquad (2)$$

where $\langle \theta, \tau \rangle = \mathbb{E}_{q_\tau}[\theta(x)]$ expresses the expected energy as a vectorized inner product. The unique maximum $\tau^*$ satisfies $q_{\tau^*}(x) = p(x; \theta)$. We call $F_{sum}(\tau, \theta) = \langle \theta, \tau \rangle + H(x; \tau)$ the sum-inference free energy (although technically the *negative* free energy).

Simply transforming a sum-inference problem into (2) does not make it easier; the marginal polytope $\mathbb{M}$ and the objective function's entropy remain intractable. However, (2) provides a framework for deriving algorithms by approximating both the marginal polytope and the entropy (Wainwright and Jordan, 2008).

Many approximation methods replace $\mathbb{M}$ with the "locally consistent" set $\mathbb{L}(G)$; in pairwise models, it is the set of singleton and pairwise beliefs $\{\tau_i | i \in V\}$ and $\{\tau_{ij} | (ij) \in E\}$ that are consistent on intersections:

$$\{\tau_i, \tau_{ij} | \sum_{x_i} \tau_{ij}(x_i, x_j) = \tau_j(x_j), \sum_{x_i} \tau_i(x_i) = 1\}.$$

Since not all such beliefs correspond to some valid joint distribution, $\mathbb{L}(G)$ is an outer bound of $\mathbb{M}$.

The free energy remains intractable (and is not even well-defined) in $\mathbb{L}(G)$. We typically approximate the free energy by a combination of singleton and pairwise entropies, which only require knowing $\tau_i$ and $\tau_{ij}$. For example, the Bethe free energy approximation (Yedidia et al., 2005) is

$$\max_{\tau \in \mathbb{L}(G)} \langle \theta, \tau \rangle + \sum_{i \in V} H_i(\tau) - \sum_{(ij) \in E} I_{ij}(\tau) \qquad (3)$$

where $H_i = -\sum_{x_i} \tau_i \log \tau_i$ is the entropy of variable $x_i$, and $I_{ij} = \sum_{x_i, x_j} \tau_{ij} \log \frac{\tau_{ij}}{\tau_i \tau_j}$ is the pairwise mutual information. Loopy BP can be interpreted as a fixed point algorithm to optimize the Bethe free energy. The tree reweighted (TRW) free energy is another variant,

$$\max_{\tau \in \mathbb{L}(G)} \langle \theta, \tau \rangle + \sum_{i \in V} H_i(\tau) - \sum_{(ij) \in E} \rho_{ij} I_{ij}(\tau) \qquad (4)$$

where $\{\rho_{ij}\}$ are edge appearance probabilities obtained from a weighted collection of spanning trees of $G$ (Wainwright et al., 2005). The TRW free energy is an upper bound of the true free energy, and is also a concave function of $\tau$ in $\mathbb{L}(G)$. Optimizing with a fixed point method gives the tree reweighted BP algorithm.

Another, related approach restricts $\mathbb{M}$ to a subset of distributions, in which both the set of constraints and the entropy calculation are tractable, such as fully factored distributions. This leads to the class of (structured) mean field approximations.

## 2.2 MAX-INFERENCE METHODS

For max-inference, we want to calculate

$$\Phi_\infty(\theta) = \max_x \{\sum_{\alpha \in \mathcal{I}} \theta_\alpha(x_\alpha)\} \qquad (5)$$

and the optimal configuration $x^*$. This problem can be shown to be equivalent to

$$\Phi_\infty(\theta) = \max_{\tau \in \mathbb{M}} \langle \theta, \tau \rangle, \qquad (6)$$

which attains its maximum when $q_\tau(x) = \delta(x = x^*)$, the Kronecker delta selecting the MAP configuration. If there are multiple MAP solutions $x^{*i}$, any convex combination $\sum_i c_i \delta(x = x^{*i})$ with $\sum_i c_i = 1, c_i \geq 0$ leads to a maximum of (6). Eqn. (6) remains NP-hard; most variational methods for MAP can be interpreted as relaxing $\mathbb{M}$ to local constraints $\mathbb{L}(G)$, which leads to a linear relaxation of the original integer programming problem. Note that (6) differs from (2) only by its lack of an entropy term; in the next section, we generalize this similarity to the marginal MAP problem.

## 3 MARGINAL MAP

Marginal MAP is simply a hybrid of the max- and sum- inference tasks. Let $A$ be a subset of nodes $V$,

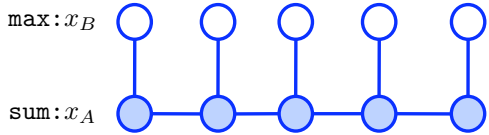

Figure 1: An example where a marginal MAP query on a tree requires exponential time. Summing over the shaded nodes makes all the unshaded nodes interdependent; see Koller and Friedman (2009) for details.

and $B = V \setminus A$ be the complement of $A$. The marginal MAP problem (mixed, or max-sum-inference) is

$$Q(x_B; \theta) = \log \sum_{x_A} \exp(\sum_{\alpha \in \mathcal{I}} \theta_\alpha(x_\alpha))$$
$$\Phi_{AB}(\theta) = \max_{x_B} Q(x_B; \theta), \qquad (7)$$

where $A$ is the set of sum nodes to be marginalized out, and $B$ is the max set, the variables to be optimized.

Although similar to max- and sum-inference, marginal MAP is significantly harder than either. A classic example in Fig. 1 shows that even on a tree, marginal MAP can be NP-hard. The main difficulty arises because the max and sum operators do not commute, which restricts efficient elimination orders to those with *all* sum nodes $x_A$ eliminated before *any* max nodes $x_B$. Marginalizing $x_A$ may destroy any conditional independence among the $x_B$, making it difficult to represent or optimize $Q(x_B; \theta)$ even if the sum part alone is tractable (such as when $A$ forms a tree). Denote $G_A = (A, E_A)$ the subgraph induced by $A$, i.e., $E_A = \{(ij) \in E | i \in A, j \in A\}$, and similarly $G_B$ and $E_B$, with $\partial_{AB} = \{(ij) \in E | i \in A, j \in B\}$ the edges that join sets $A$ and $B$. The natural generalization of an efficient tree structure to the marginal MAP problem occurs when $G$ is a tree along an elimination order that first eliminates all nodes in $A$, then those in $B$. We call this type of graph an *A-B tree*.

For these reasons, there are relatively few algorithms (particularly approximate algorithms) for marginal MAP. Expectation-maximization (EM) or variational EM provide one straightforward approach, by viewing $x_B$ as the parameters and $x_A$ as the hidden variables; however, EM has many local maxima and is easily stuck at sub-optimal configurations. Jiang and Daumé III (2010) proposed a message passing algorithm combining max-product and sum-product BP, but with little theoretical analysis. Other state-of-the-art approaches include Markov chain Monte Carlo (Doucet et al., 2002) and local search (Park and Darwiche, 2004). In this work, we propose a general variational framework for approximation algorithms of marginal MAP, and provide both theoretical and experimental results to justify our algorithms.

## 4 A VARIATIONAL APPROACH

As one main result of this work, in this section we derive a dual representation of the marginal MAP problem (7). The dual form generalizes that of sum-inference in (2) and max-inference in (6), and provides a unified framework for addressing marginal MAP.

**Theorem 4.1.** $\Phi_{AB}$ *has a dual representation*

$$\Phi_{AB}(\theta) = \max_{\tau \in \mathbb{M}} \{\langle \theta, \tau \rangle + H(x_A|x_B; \tau)\}, \qquad (8)$$

*where $\mathbb{M}$ is the marginal polytope; $H(x_A|x_B) = -\sum_x q_\tau(x) \log q_\tau(x_A|x_B)$ is the conditional entropy, with $q_\tau(x)$ being the maximum entropy distribution corresponding to $\tau$.[1] If $Q(x_B; \theta)$ has a unique maximum $x_B^*$, the maximum $\tau^*$ of (8) is also unique, with $q_\tau^*(x_B) = \delta(x_B = x_B^*)$ and $q_\tau^*(x_A|x_B^*) = p(x_A|x_B^*; \theta)$; if there are multiple (global) maxima $x_B^{*i}$, $q_\tau^*(x_B)$ can be any convex combination of these optima, i.e., $q_\tau^*(x_B) = \sum_i c_i \delta(x_B = x_B^{*i})$ with $\sum_i c_i = 1$ and $c_i \geq 0$.*

*Proof.* For an arbitrary distribution $q(x)$, consider the conditional KL divergence

$$\mathbb{E}_q[D(q(x_A|x_B)||p(x_A|x_B; \theta))] = \sum_x q(x) \log \frac{q(x_A|x_B)}{p(x_A|x_B)}$$
$$= -H(x_A|x_B; q) + \mathbb{E}_q[\log p(x_A|x_B)]$$
$$= -H(x_A|x_B; q) + \mathbb{E}_q[\theta(x)] - \mathbb{E}_q[Q(x_B; \theta)] \geq 0.$$

where the last inequality follows from the nonnegativity of KL divergence, and is tight iff $q(x_A|x_B) = p(x_A|x_B; \theta)$ for all $x_A$ and $x_B$ that $q(x_B) \neq 0$. Therefore, we have

$$\Phi_{AB}(\theta) \geq \mathbb{E}_q[Q(x_B; \theta)] \geq \mathbb{E}_q[\theta(x)] + H(x_A|x_B; q).$$

It is easy to show that the two inequality signs are tight if and only if $q(x)$ equals $q_\tau^*(x)$ as defined above. Substituting $\mathbb{E}_q[\theta(x)] = \langle \theta, \tau \rangle$ completes the proof. □

Note that since $H(x_A|x_B) = H(x) - H(x_B)$, Theorem 4.1 transforms the marginal MAP problem into the maximization of a "truncated" free energy

$$F_{mix}(\tau, \theta) = \langle \theta, \tau \rangle + H(x_A|x_B) = F_{sum}(\tau, \theta) - H(x_B)$$

where the entropy $H(x_B)$ of the max nodes $x_B$ are removed from the sum-inference free energy $F_{sum}$. This generalizes sum-inference (2) and max-inference (6), where the max sets are empty and all nodes respectively. Intuitively, by subtracting the entropy $H(x_B)$ in the objective, the marginal $q_\tau(x_B)$ tends to have lower entropy, causing its probability mass to concentrate on the optimal set $\{x_B^{*i}\}$.

---

[1] Although the optimal $\tau$ has some zero entries, the maximum entropy distribution remains unique (Jaynes, 1957).

The optimal $\tau^*$ of (8) can be interpreted as corresponding to a distribution obtained by clamping the value of $x_B$ at the optimal $B$-configuration $x_B^*$ on the distribution $p(x;\theta)$, i.e., $q_{\tau^*}(x) = p(x|x_B = x_B^*;\theta)$.

On the other hand, for any marginal MAP solution $x_B^{*i}$, the $\tau^{*i}$ with $q_\tau^{*i}(x_B) = p(x|x_B = x_B^{*i};\theta)$ is also an optimum of (8). Therefore, the optimization in (8) can be restricted to $\mathbb{M}^* = \{\tau | q_\tau(x) = p(x|x_B = x_B^*;\theta), \forall x_B^* \in \mathcal{X}^{|B|}$ and $\theta = \sum_{\alpha \in I} \theta_\alpha\}$, corresponding to the set of distributions in which $x_B$ are clamped to some value. That is, we also have $\Phi_{AB} = \max_{\tau \in \mathbb{M}^*} F_{mix}(\tau, \theta)$. More generally, the same holds for any set $\mathbb{N}$ that satisfies $\mathbb{M}^* \subset \mathbb{N} \subset \mathbb{M}$ without affecting the optimum. Among these sets, $\mathbb{M}$ is of special interest because it is the smallest convex set that includes $\mathbb{M}^*$, i.e., it is the convex hull of $\mathbb{M}^*$.

Theorem 4.1 transforms the marginal MAP problem into a variational form, but does not decrease the hardness – both the marginal polytope $\mathbb{M}$ and the free energy $F_{mix}(\tau, \theta)$ remain intractable. Fortunately, the well established techniques for sum- and max-inference can be directly applied to (8), giving a new way to derive approximate algorithms. In the spirit of Wainwright and Jordan (2008), one can either relax $\mathbb{M}$ to a simpler outer bound like $\mathbb{L}(G)$, and replace $F_{mix}(\tau, \theta)$ by some tractable form to give algorithms similar to LBP or TRBP, or restrict $\mathbb{M}$ to a subset in which constraints and free energy are tractable to give a mean field-like algorithm. In the sequel, we introduce several such approximation schemes. We mainly focus on BP analogues, although we briefly discuss mean field when we connect to EM in section 7.

**Bethe-like free energy.** Motivated by the regular Bethe approximation (3), we approximate the marginal MAP dual (8) by

$$\Phi_{bethe}(\theta) = \max_{\tau \in \mathbb{L}(G)} F_{bethe}(\tau, \theta), \quad (9)$$

$$F_{bethe}(\tau, \theta) = \langle \theta, \tau \rangle + \sum_{i \in A} H_i - \sum_{(ij) \in E_A \cup \partial_{AB}} I_{ij}. \quad (10)$$

where we call $F_{bethe}$ a "truncated" Bethe free energy, since it can be obtained from the regular sum-inference Bethe free energy by truncating (discarding) the entropy and mutual information terms that involve only max nodes. If $G$ is a $A$-$B$ tree, $\Phi_{bethe}$ equals the true $\Phi_{AB}$, giving an intuitive justification. In the sequel we give more general conditions under which this approximation can give the exact solution. We find that this simple scheme can usually give high quality empirical approximations. Similar to the regular Bethe approximation, (9) leads to a nonconvex optimization, and we can derive both message passing algorithms and provably convergent algorithms to solve it.

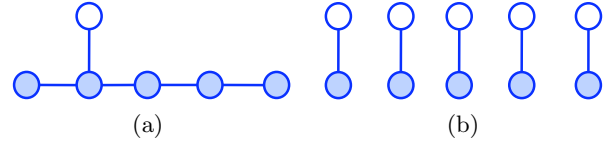

Figure 2: (a) A type-I $A$-$B$ subtree and (b) a type-II $A$-$B$ subtree of the hidden Markov chain in Fig. 1.

**Tree-reweighted free energy.** Following TRW, we construct an approximation of marginal MAP using a convex combination of $A$-$B$ subtrees. Suppose $\mathcal{T}_{AB} = \{T\}$ is a collection of $A$-$B$ subtrees of $G$, where each $T$ is assigned a weight $w_T$ with $w_T \geq 0$ and $\sum_{T \in \mathcal{T}} w_T = 1$. For each $A$-$B$ tree $T = (V, E_T)$, define

$$H_T(x_A|x_B; \tau) = \sum_{i \in A} H_i(\tau) - \sum_{(ij) \in E_T \setminus E_B} I_{ij}(\tau);$$

this is always a concave function of $\tau \in \mathbb{L}(G)$, and $H(x_A|x_B; \tau) \leq H_T(x_A|x_B; \tau)$. More generally, we have $H(x_A|x_B) \leq \sum_{T \in \mathcal{T}} w_T H_T(x_A|x_B)$ which can be transformed to

$$F_{trw}(\theta, \tau) = \langle \theta, \tau \rangle + \sum_{i \in A} H_i - \sum_{(ij) \in E_A \cup \partial_{AB}} \rho_{ij} I_{ij}, \quad (11)$$

where $\rho_{ij} = \sum_{T:(ij) \in E_T} w_T$ are the usual edge appearance probabilities. Replacing $\mathbb{M}$ with $\mathbb{L}(G)$ and $F$ with $F_{trw}$ leads to an approximation of (8)

$$\Phi_{trw}(\theta) = \max_{\tau \in \mathbb{L}(G)} F_{trw}(\tau, \theta). \quad (12)$$

Since $\mathbb{L}(G)$ is an outer bound of $\mathbb{M}$, and $F_{trw}$ is a concave upper bound of the true free energy, we can guarantee that $\Phi_{trw}$ is always an upper bound of $\Phi_{AB}$, to our knowledge the first known upper bound for marginal MAP.

**Selecting $A$-$B$ subtrees.** Selecting $A$-$B$ subtrees for approximation (12) is not as straightforward as selecting subtrees in regular sum-inference. An important property of an $A$-$B$ tree $T$ is that no two edges of $T$ in $\partial_{AB}$ can be connected by edges or nodes of $T$ in $G_A$. Therefore, one can construct an $A$-$B$ subtree by first selecting a subtree in $G_A$, and then join each connected component of $G_A$ to at most one edge in $\partial_{AB}$. Two simple, extreme cases stand out:

(i) *type-I $A$-$B$ subtrees*, which include a spanning tree of $G_A$ and only one crossing edge in $\partial_{AB}$;
(ii) *type-II $A$-$B$ subtrees*, which include no edges in $G_A$, but several edges in $\partial_{AB}$ that are not incident on the same nodes in $G_A$.

See Fig. 2 for an example. Intuitively, type-I subtrees capture more information about the summation structures of $G_A$, while type-II subtrees capture more information about $\partial_{AB}$, relating the sum and max parts.

If one restricts to the set of type-I subtrees, it is possible to guarantee that, if $G_A$ is a tree, the summation component will be exact (all $\rho_{ij} = 1$ for $(ij) \in E_A$), in which case it will be possible to make some theoretical guarantees about the solution. However in experiments we find it is often practically beneficial to balance type-I and type-II when choosing the weights.

**Global Optimality.** It turns out that the above approximation schemes can give exact solutions under some circumstances. We initially assume that $G_A$ is tree, i.e., the sum part is tractable to calculate for a given $B$-configuration. (Note that the marginal MAP problem remains hard even in this case, as suggested by Fig. 1.) Suppose we approximate $\Phi_{AB}(\theta)$ by

$$\Phi_{tree} = \max_{\tau \in \mathbb{L}(G)} \langle \theta, \tau \rangle + \sum_{i \in V} H_i - \sum_{(ij) \in E_A} I_{ij} - \sum_{(ij) \in \partial_{AB}} \rho_{ij} I_{ij} \quad (13)$$

where $\{\rho_{ij} | (ij) \in \partial_{AB}\}$ can take arbitrary values, while $\{\rho_{ij} | (ij) \in E_A\}$ have been fixed to be ones; this assumption guarantees that the sum part is "intact" in the approximation. Finally, we assume that (13) is *globally* optimized.

**Theorem 4.2.** *Suppose $G_A$ is a tree, and we approximate $\Phi_{AB}(\theta)$ using $\Phi_{tree}$ defined in (13). We have*

(i) $\Phi_{tree}(\theta) \geq \Phi_{AB}(\theta)$. *If the there exist $x_B^*$ such that $Q(x_B^*; \theta) = \Phi_{tree}(\theta)$, we have $\Phi_{tree}(\theta) = \Phi_{AB}(\theta)$, and $x_B^*$ is an optimal marginal MAP solution.*

(ii) *Suppose $\tau^*$ is a global maximum of (13), and $\{\tau_i^*(x_i) | i \in B\}$ are integral, i.e., $\tau_i^*(x_i) = 0$ or $1$, then $\{x_i^* = \arg\max_{x_i} \tau_i^*(x_i) | i \in B\}$ is an optimal solution of the marginal MAP problem (7).*

*Proof.* As discussed in Section 4, the optimization in (8) can be restricted on $\mathbb{M}^*$, the subset of $\mathbb{M}$ in which $\tau(x_B)$ are integral, that is, $\Phi_{AB} = \max_{\tau \in \mathbb{M}^*} F_{mix}$. Note that the objective function in (13) equals the true free energy $F_{mix}(\tau, \theta)$ when $\tau \in \mathbb{M}_B^*$ and $G_A$ is a tree (since $I_{ij} = 0$, $(ij) \in \partial_{AB}$, for $\forall \tau \in \mathbb{M}_B^*$). This means that (13) is a relaxation of $\Phi_{AB} = \max_{\tau \in \mathbb{M}^*} F_{mix}$. A standard relaxation argument completes the proof. □

Theorem 4.2 gives some justification for both the Bethe free energy, in which $\rho_{ij} = 1$ for all $(ij) \in \partial_{AB}$, and the TRW free energy with only type-I subtrees, in which $\sum_i \rho_{ij} = 1$, $\rho_{ij} \geq 0$. However, it may have limited practical application. On the one hand, the Bethe free energy is usually non-concave, and it is hard to show that a solution is globally optimal. The concavity is controlled by the value of $\{\rho_{ij} | (ij) \in \partial_{AB}\}$; small enough $\rho_{ij}$ (as in TRW) guarantees concavity. On the other hand, the values of $\{\rho_{ij} | (ij) \in \partial_{AB}\}$ also control how likely the solution is to be integral – larger $\rho_{ij}$ emphasizes the mutual information terms, forcing the solution towards integral points. Thus in practice the solution of the TRW free energy is less likely to be integral than the Bethe free energy, causing difficulty in applying Theorem 4.2 to TRW solutions as well. In general, the values of $\{\rho_{ij} | (ij) \in \partial_{AB}\}$ reflect a tradeoff between concavity and integrality. Interestingly, as we will show later, the EM algorithm can be also viewed as optimizing an objective of the form (13) by setting $\rho_{ij} \to +\infty$, which strongly forces solution integrality, but causes a highly non-convex objective. It appears that by setting $\rho_{ij} = 1$, the Bethe free energy obtains a good tradeoff, giving excellent performance.

We give a more practical statement of optimality in Section 5.1, related to local optima.

## 5 MIXED MESSAGE PASSING

We see that the general objective function

$$\max_{\tau \in \mathbb{L}(G)} \langle \theta, \tau \rangle + \sum_{i \in V} w_i H_i - \sum_{(ij) \in E} w_{ij} I_{ij}, \quad (14)$$

can be used to approximate sum-inference, max-inference and mixed-inference problems simply by taking different weights $w_i$, $w_{ij}$. If $w_i = 1$ for $\forall i \in V$, (14) addresses the sum-inference problem, and loopy BP and its variants can be derived as fixed point optimizers of (14). If $w_i = 0$ for $\forall i \in V$, (14) addresses the max-inference problem, and max-product BP variants can be derived as a zero-temperature limit of sum-product BP, as $w_i$ and $w_{ij}$ approach zero. We have shown that mixed-inference lies between the two ends of the spectrum – as $w_i = 0$ only for $\forall i \in B$, (14) addresses the mixed-inference problem. Given these connections, we can expect some "mixed" message passing algorithms for marginal MAP that combine max-product BP and sum-product BP by allowing weights in the `max` set to approach zero, while keeping weights in the `sum` set equal to one. In this section, we derive such a "mixed" message scheme, and discuss an optimality property of its fixed points using a reparameterization interpretation.

To start, consider the case when $w_i$ and $w_{ij}$ are strictly positive. Using a Lagrange multiplier method similar to Yedidia et al. (2005) or Wainwright et al. (2005), we can show that the fixed point of the following message passing scheme is a stationary point of (14):

$$m_{i \to j}(x_j) \leftarrow \Big[ \sum_{x_i} (\psi_i m_{\sim i})^{1/w_i} (\psi_{ij}/m_{j \to i})^{1/w_{ij}} \Big]^{w_{ij}}$$

$$\tau_i(x_i) \propto (\psi_i m_{\sim i})^{1/w_i} \quad (15)$$

$$\tau_{ij}(x_{ij}) \propto \tau_i \tau_j \Big( \frac{\psi_{ij}}{m_{i \to j} m_{j \to i}} \Big)^{1/w_{ij}}$$

where $m_{\sim i} = \prod_{k \in \mathcal{N}(i)} m_{k \to i}$ is the product of the messages sent to $i$. Unfortunately, if some weights are zero the message passing algorithm can not be derived directly, mainly because the inequality constraints $\tau_{ij} \geq 0$ (which can be ignored for strictly positive weights) must be explicitly considered. (For detailed discussion of this issue, see Wainwright and Jordan (2008) and Yedidia et al. (2005)). However, we can apply (15) on positive weights that are close to zero, and hope the solution is close enough to the marginal MAP solution. Let $\hat{F}(\tau, \theta)$ be a surrogate free energy of marginal MAP as defined in (11) and (10) (e.g., $\hat{F}$ can be either $F_{trw}$ or $F_{bethe}$). Let

$$\hat{H}(x_B; \tau) = \sum_{i \in B} H_i(\tau) - \sum_{(ij) \in E_B} \rho_{ij} I_{ij}(\tau),$$

where $\rho_{ij} > 0$ for $(ij) \in E_B$. For $\epsilon > 0$, define

$$\hat{F}^\epsilon(\tau, \theta) = \hat{F}(\tau, \theta) + \epsilon \hat{H}(x_B; \tau)$$
$$= \langle \theta, \tau \rangle + \sum_{i \in V} w_i(\epsilon) H_i - \sum_{(ij) \in E} w_{ij}(\epsilon) I_{ij}$$

where $w_i(\epsilon) = \epsilon$, $w_{ij}(\epsilon) = \epsilon \rho_{ij}$ for $i \in B, (ij) \in E_B$ and $w_i(\epsilon) = 1$, $w_{ij}(\epsilon) = \rho_{ij}$ otherwise. We can see that $w_i(\epsilon)$ and $w_{ij}(\epsilon)$ are positive by definition, and can therefore solve $\tau^*(\epsilon) = \arg\max_{\tau \in \mathbb{L}(G)} \{\hat{F}^\epsilon\}$ using message update (15) for small $\epsilon > 0$ and hope that $\tau^*(\epsilon)$ approaches the solution $\tau^*$ of $\hat{F}$ as $\epsilon \to 0$.

Unfortunately, this is not always true. Weiss et al. (2007) showed that for max-inference (i.e., when $\hat{F} = \langle \theta, \tau \rangle$), $\tau^*(\epsilon)$ approaches $\tau^*$ only when the augmented term $\hat{H}(x_B)$ is concave. We generalize this result to arbitrary $\hat{F}$, and give an error bound for concave $\hat{F}$.

We say that $\hat{H}(x_B; \tau)$ is provably concave if it can be reformed into $\hat{H}(x_B; \tau) = \sum_{i \in B} \kappa_i H(x_i) + \sum_{(ij) \in E_B} \kappa_{ij} H(x_i|x_j)$ for some positive $\kappa_i$ and $\kappa_{ij}$. This is equivalent to saying $\kappa_i + \sum_{j \in \mathcal{N}(i)} \kappa_{ij} = 1$, $\kappa_{ij} + \kappa_{ji} = \rho_{ij}$. Following Weiss et al. (2007), we call such set of weights $\{\rho_{ij}|(ij) \in E_B\}$ "provably concave". We can establish the following result.

**Theorem 5.1.** Let $\tau^*$ be a stationary point of $\hat{F}(\tau, \theta)$ in $\mathbb{L}(G)$, and $\tau^*(\epsilon)$ a stationary point of $\hat{F}^\epsilon(\tau, \theta) = \hat{F}(\tau, \theta) + \epsilon \hat{H}(x_B; \tau)$ in $\mathbb{L}(G)$. If $\hat{H}(x_B; \tau)$ is provably concave, we have

(i) Let $\{\epsilon_k\}$ be a sequence of positive numbers that approaches zero and $\tau^*(\infty)$ be a limit point of $\{\tau^*(\epsilon_k)|k = 1, 2, \cdots\}$; then $\tau^*(\infty)$ is a stationary point of $\hat{F}(\tau, \theta)$ in $\mathbb{L}(G)$ (regardless of whether $\hat{F}(\tau, \theta)$ is concave).

(ii) Further, if $\hat{F}(\tau, \theta)$ is a concave function of $\tau$ in $\mathbb{L}(G)$, we can give an error bound

$$0 \leq \hat{F}(\tau^*, \theta) - \hat{F}(\tau^*(\epsilon), \theta) \leq \epsilon \hat{H}(x_B; \tau^*(\epsilon)) \leq \epsilon |B| \log |\mathcal{X}|$$

where $|\mathcal{X}|$ is the number of states that $x_i$ can take ($|\mathcal{X}| = 2$ for binary variables), and $|B|$ is the number of nodes in $B$.

*Proof.* The proof of (i) involves showing $\tau^*(\infty)$ satisfies the KKT condition of $\max_{\tau \in \mathbb{L}(G)} \hat{F}$. The second inequality in (ii) follows by showing that $\hat{F}^\epsilon(\tau^*(\epsilon), \theta)$ is a point on the dual function of $\hat{F}$, which gives an upper bound of $\hat{F}$; the third inequality in (ii) follows from the fact that $H(x_i) \leq \log |\mathcal{X}|$ and $I_{ij} \geq 0$. □

In practice, there is usually no reason to run the algorithm with very small $\epsilon$; we instead directly take the limit on the message passing scheme, using

$$\lim_{w \to 0^+} [\sum_x f(x)^{1/w}]^w = \max_x f(x)$$

where $f(x)$ is an arbitrary positive function and $w \to 0^+$ represents $w$ approaching 0 from the positive side. We can show that the message scheme in (15) with $w_i(\epsilon)$, $w_{ij}(\epsilon)$ then approaches a "mixed" message scheme as $\epsilon \to 0$, which depends on the node type of source and destination:

$$A \to V: \quad m_{i \to j} \leftarrow \Big[\sum_{x_i}(\psi_i m_{\sim i})(\frac{\psi_{ij}}{m_{j \to i}})^{1/\rho_{ij}}\Big]^{\rho_{ij}}$$

$$B \to B: \quad m_{i \to j} \leftarrow \max_{x_i}(\psi_i m_{\sim i})^{\rho_{ij}}(\frac{\psi_{ij}}{m_{j \to i}}) \quad (16)$$

$$B \to A: \quad m_{i \to j} \leftarrow \Big[\sum_{x_i \in \arg\max\{\psi_i m_{\sim i}\}}(\frac{\psi_{ij}}{m_{j \to i}})^{1/\rho_{ij}}\Big]^{\rho_{ij}}$$

where $V = A \cup B$ are all nodes. Sum-product messages are sent from `sum` nodes to any other nodes; max-product messages are sent between `max` nodes; and the messages sent from `max` nodes to `sum` nodes are novel and will be interpreted in the sequel as solving a type of local marginal MAP problem.

Interestingly, our method bears similarity to, but has key differences from the recent method of Jiang and Daumé III (2010), who propose a similar hybrid message passing algorithm, but that sends the usual max-product messages from `max` to `sum` nodes. It turns out that this difference is crucial for the analysis of optimality conditions, as we discuss later.

We define a set of "mixed-marginals" as:

$$b_i(x_i) \propto (\psi_i m_{\sim i})$$
$$b_{ij}(x_{ij}) \propto b_i b_j (\frac{\psi_{ij}}{m_{i \to j} m_{j \to i}})^{1/\rho_{ij}} \quad (17)$$

The maximum of the mixed-marginals $x_i^* \in \arg\max_{x_i} b_i(x_i), \forall i \in B$ can be extracted as an estimate of the marginal MAP solution, as is typical in max-product BP. We will prove that such $x_B^*$ are locally optimal under some conditions.

The mixed-marginals are not expected to approach the optimum $\tau^*$, but are rather softened versions of $\tau^*$. The following theorem clarifies their relationship:

**Theorem 5.2.** *Let $\{\tau_{i,\epsilon}, \tau_{ij,\epsilon}\}$ be a fixed point of weighted message passing (15) under weights $\{w_i(\epsilon), w_{ij}(\epsilon)\}$. Define*

$$\begin{cases} b_{i,\epsilon} = (\tau_{i,\epsilon})^\epsilon & \forall i \in B \\ b_{i,\epsilon} = \tau_{i,\epsilon} & \forall i \in A \\ b_{ij,\epsilon} = b_{i,\epsilon} b_{j,\epsilon} (\frac{\tau_{ij,\epsilon}}{\tau_{i,\epsilon} \tau_{j,\epsilon}})^\epsilon & \forall (ij) \in E_B \\ b_{ij}^\epsilon = b_{i,\epsilon} b_{j,\epsilon} (\frac{\tau_{ij,\epsilon}}{\tau_{i,\epsilon} \tau_{j,\epsilon}}) & \forall (ij) \in E_A \cup \partial_{AB}, \end{cases}$$

*then $\{b_{i,\epsilon}, b_{ij,\epsilon}\}$ approaches the mixed-marginals (17) of the fixed point of the mixed message scheme (16).*

*Proof.* The result follows directly from application of the zero temperature limit; see Section 4 in Weiss et al. (2007) for a similar proof for max-product. □

### 5.1 LOCAL OPTIMALITY VIA REPARAMETERIZATION

An important interpretation of the sum-product and max-product algorithms is the reparameterization viewpoint (Wainwright et al., 2003; Weiss et al., 2007). Message passing can be viewed as moving mass between the sum-marginals (resp. max-marginals), in a way that leaves their product a reparameterization of the original distribution; at any fixed point, the sum (resp. max) marginals are guaranteed to satisfy the sum (resp. max) consistency property.

Interestingly, the mixed-marginals have a similar reparameterization interpretation.

**Theorem 5.3.** *The mixed-marginals (17) at the fixed point of the mixed message scheme (16) satisfies*

***Admissibility:***

$$p(x) \propto \prod_{i \in V} b_i(x_i) \prod_{(ij) \in E} \Big[\frac{b_{ij}(x_i, x_j)}{b_i(x_i) b_j(x_j)}\Big]^{\rho_{ij}}$$

***Mixed-consistency:***

(a) $\sum_{x_i} b_{ij}(x_i, x_j) = b_j(x_j), \quad \forall i \in A, j \in A \cup B$

(b) $\max_{x_i} b_{ij}(x_i, x_j) = b_j(x_j), \quad \forall i \in B, j \in B$

(c) $\sum_{x_i \in \arg\max b_i} b_{ij}(x_i, x_j) = b_j(x_j), \quad \forall i \in B, j \in A$

*Proof.* Directly substitute (16) into (17). □

The three mixed-consistency constraints exactly map to the three types of message updates in (16). Constraint (c) is of particular interest: it can be interpreted as solving some local marginal-MAP problem $x_i = \arg\max \sum_{x_j} b_{ij}$. It turns out that this constraint is a crucial ingredient of mixed message passing, enabling us to prove local optimality of a solution.

Suppose $C$ is a subset of max nodes in $B$, $G_{C \cup A} = (C \cup A, E_{C \cup A})$ is the subgraph of $G$ induced by nodes $C \cup A$, where $E_{C \cup A} = \{(ij) \in E | i, j \in C \cup A\}$. We call $G_{C \cup A}$ a semi-$A$-$B$ subtree if the edges in $E_{C \cup A} \backslash E_B$ form an $A$-$B$ tree. In other words, $G_{C \cup A}$ is a semi-$A$-$B$ tree if it is an $A$-$B$ tree when ignoring the edges within max set.

**Theorem 5.4.** *Let $\{b_i, b_{ij}\}$ be a set of mixed-marginals that satisfy the admissibility and mixed-consistency of Theorem 5.3. Suppose the maxima of $b_i$, $b_{ij}$ are all unique. Then there exist a $B$-configuration $x_B^*$ satisfying $x_i^* \in \arg\max b_i$ for $\forall i \in B$ and $(x_i^*, x_j^*) \in \arg\max b_{ij}$ for $\forall (ij) \in E_B$. Suppose $C$ is a subset of $B$ such that $G_{C \cup A}$ is a semi-$A$-$B$ tree, we have that $x_B^*$ is locally optimal in the sense that $Q(x_B^*; \theta)$ is not smaller than any $B$-configuration that differs from $x_B^*$ only on $C$, if all the following conditions are satisfied:*

(i) $\rho_{ij} = 1$ for $(ij) \in E_A$.
(ii) $0 \leq \rho_{ij} \leq 1$ for $(ij) \in E_{C \cup A} \cap \partial_{AB}$.
(iii) $\{\rho_{ij} | (ij) \in E_{C \cup A} \cap E_B\}$ *is provably concave.*

*Proof.* Since the maximum of $b_i$, $b_{ij}$ are unique, we have $b_{ij}(x_i, x_j^*) = b_i(x_i)$ for $i \in A$, $j \in B$. The fact that $G_{C \cup A}$ is a semi-$A$-$B$ tree enables the summation part to be eliminated away. The remaining part only involves the max nodes, and the analysis of Weiss et al. (2007) applies. □

For $G_{C \cup A}$ to be a tree, the sum graph $G_A$ must be a tree. Thus, Theorem 5.4 implicitly assumes that the sum part is tractable. For the Markov chain in Fig. 1, Theorem 5.4 implies only that the solution is locally optimal up to Hamming distance one, i.e., coordinate-wise optimal. However, the local optimality guaranteed by Theorem 5.4 is in general much stronger when the sum part is disconnected, or the max part has interior regions that do not connect to the sum part.

We emphasize that mixed-consistency constraint (c) plays an important role in the proof of Theorem 5.4, canceling the terms that involve variables in $B \backslash C$. The hybrid algorithm in Jiang and Daumé III (2010) also has an reparameterization interpretation, which replaces our constraint (c) with simple max-consistency; however this change invalidates Theorem 5.4.

## 6 CONVERGENT ALGORITHMS BY CCCP

The mixed message scheme (16) is interesting, but may suffer from convergence problems, as can happen to loopy BP or tree reweighted BP. We apply a concave-convex procedure (CCCP) used to derive convergent algorithms for maximizing the Bethe and Kikuchi free energy (Yuille, 2002) to our problem.

Suppose $\hat{F}(\tau, \theta) = \langle \theta, \tau \rangle + \hat{H}$ is our surrogate free energy, where $\hat{H}$ is the approximation of the conditional entropy term. Let us decompose the entropy term into positive and negative part $\hat{H} = \hat{H}^+ - \hat{H}^-$, where

$$\hat{H}^+ = \sum_{i \in V} \omega_i^+ H_i - \sum_{(ij) \in E} \omega_{ij}^+ I_{ij},$$
$$\hat{H}^- = \sum_{i \in V} \omega_i^- H_i - \sum_{(ij) \in E} \omega_{ij}^- I_{ij}. \quad (18)$$

Suppose $\max_{\tau \in \mathbb{L}(G)} \langle \theta, \tau \rangle + H^+$ is easy to solve. We can optimize the marginal MAP free energy by iteratively linearizing the $H^-$ term, giving

$$\theta_{ij}^{n+1} = \theta_{ij}^n + \omega_{ij}^- \log \frac{\tau_{ij}^n}{\tau_i^n \tau_j^n}, \quad \theta_i^{n+1} = \theta_i^n + \omega_i^- \log \tau_i^n$$
$$\tau^{n+1} \leftarrow \arg \max_{\tau \in \mathbb{L}(G)} \langle \theta^{n+1}, \tau \rangle + H^+. \quad (19)$$

where we use the fact that the gradient of $H_i$ and $I_{ij}$ w.r.t. $\tau_i$ and $\tau_{ij}$ are $-\log \tau_i$ and $\log \frac{\tau_{ij}}{\tau_i \tau_j}$ respectively. If both $H^+$ and $H^-$ are concave, the iterative linearization process (19) is called the concave-convex procedure (CCCP), and it is guaranteed to monotonically increase the objective function. See Yuille (2002) for a detailed discussion.

This iterative linearization process has an appealing interpretation. Recall that the free energy of marginal MAP is obtained by dropping the entropy of the max nodes from the sum-inference free energy. Eq. (19) essentially "adds back" the lost entropy terms, while canceling their effect by adjusting $\theta$ in the opposite direction. This concept is distinct from the technique we used when deriving mixed message passing, in which a truncated entropy term was re-added but weighted with a $\epsilon$-small temperature.

In practice, we does not necessary require $H^+$ and $H^-$ to be concave; in particular, it may be appealing to choose $\langle \theta, \tau \rangle + H^+$ to coincide with the Bethe free energy for sum-inference when using the truncated Bethe approximation. This has the interpretation of transforming the marginal MAP problem into a sequence of sum-inference problems, and often appears to give a better fixed point solution.

## 7 CONNECTIONS TO EM

A natural algorithm for solving the marginal MAP problem is to use the expectation-maximization (EM) algorithm, by treating $x_B$ as the parameters and $x_A$ as the hidden variables. In this section, we show that the EM algorithm can be seen as a coordinate ascent algorithm on a mean variant of our framework.

To connect to EM, let us restrict $\mathbb{M}$ to $\mathbb{M}^\times$, set of distributions with a product form on pairs $(x_A, x_B)$,

i.e., $\mathbb{M}^\times = \{\tau \in \mathbb{M} | q_\tau(x) = q_\tau(x_A) q_\tau(x_B)\}$. Since $\mathbb{M}^* \subset \mathbb{M}^\times \subset \mathbb{M}$, meaning that the set of optimal vertices are included in $\mathbb{M}^\times$, $\max_{\tau \in \mathbb{M}^\times} F_{mix}(\tau, \theta)$ remains exact; however, $\mathbb{M}^\times$ is no longer a convex set.

Denoting $\mathbb{M}_A$ as the marginal polytope over $x_A$, and similarly for $\mathbb{M}_B$, it is natural to consider a coordinate update for the restricted optimization:

$$\text{Sum}: \quad \tau_A^{n+1} \leftarrow \underset{\tau_A \in \mathbb{M}_A}{\operatorname{argmax}} \langle \mathbb{E}_{q_B^n}(\theta), \tau_A \rangle + H_{\tau_A}(x_A)$$
$$\text{Max}: \quad \tau_B^{n+1} \leftarrow \underset{\tau_B \in \mathbb{M}_B}{\operatorname{argmax}} \langle \mathbb{E}_{q_A^{n+1}}(\theta), \tau_B \rangle \quad (20)$$

where $q_A^n(x_A)$ and $q_B^n(x_B)$ are the maximum entropy distribution of $\tau_A^n \in \mathbb{M}_A$ and $\tau_B^n \in \mathbb{M}_B$. Note that the sum and max step each happen to be the dual of a sum-inference and max-inference problem respectively. If we go back to the primal, and update the primal configuration $x_B$ instead of $\tau_B$, (20) can be rewritten

$$\text{E step}: \quad q_A^{n+1}(x_A) \leftarrow p(x_A | x_B^n; \theta)$$
$$\text{M step}: \quad x_B^{n+1} \leftarrow \arg \max_{x_B} \mathbb{E}_{\tau_A^{n+1}}(\theta), \quad (21)$$

which is an EM update viewing $x_B$ as parameters and $x_A$ as hidden variables. EM is also connected to coordinate ascent on variational objectives in Neal and Hinton (1998) and Wainwright and Jordan (2008).

When the E-step or M-step are intractable, one can insert various approximations. In particular, approximating $\mathbb{M}_A$ by a mean-field inner bound $\mathbb{M}_A^{mf}$ leads to variational EM. An interesting observation is obtained by using Bethe approximation (3) to solve the E-step and linear relaxation to solve the M-step; in this case, the EM-like update is equivalent to solving

$$\max_{\tau \in \mathbb{L}^\times(G)} \langle \theta, \tau \rangle + \sum_{i \in A} H_i - \sum_{(ij) \in E_A} I_{ij} \quad (22)$$

where $\mathbb{L}^\times(G)$ is the subset of $\mathbb{L}(G)$ in which $\tau_{ij}(x_i, x_j) = \tau_i(x_i) \tau_j(x_j)$ for $(ij) \in \partial_{AB}$. Equivalently, $\mathbb{L}^\times(G)$ is the subset of $\mathbb{L}(G)$ in which $I_{ij} = 0$ for $(ij) \in \partial_{AB}$. Therefore, (22) can be treated as an special case of (13) by taking $\rho_{ij} \to +\infty$, forcing the solution $\tau^*$ to fall into $\mathbb{L}^\times(G)$. The EM algorithm represents an extreme of this tradeoff, encouraging vertex solutions by sacrificing convexity. The optimality result in Theorem 4.2 also applies to EM, but EM solutions are likely to be stuck in local optima due to the high non-convexity.

## 8 EXPERIMENTS

We first illustrate our algorithm on the hidden Markov chain example shown in Fig. 1, and then show its behavior on a harder, Ising-like model.

**Hidden Markov chain.** We define a distribution on the HMM in Fig. 1 with 20 nodes by

$$p(x) \propto \exp\big[\sum_i \theta_i(x_i) + \sum_{(ij) \in E} \theta_{ij}(x_i, x_j)\big]$$

with $x \in \{-1, 0, +1\}$. We set $\theta_{ij}(k, k) = 0$ for all $(ij)$ and $k$, and randomly generate $\theta_i(k) \sim \mathcal{N}(0, 0.1)$, $\theta_{ij}(k, l) \sim \mathcal{N}(0, \sigma)$ for $k \neq l$, where $\sigma$ is a coupling strength. Our results are averaged over 100 sets of random parameters for each $\sigma \in [0, 1.5]$.

We implemented the truncated Bethe approximation (`Mix-Bethe`) and two versions of truncated TRW approximations: `Mix-TRW1` which assigns all weights uniformly on the set of type-I subtrees, and `Mix-TRW2` which assigns weight 0.5 uniformly over type-I subtrees, and weight 0.5 uniformly over type-II trees. To avoid convergence issues, we use CCCP for all our algorithms. For comparison, we implemented max-product BP (`Max-product`), sum-product BP (`Sum-product`) and Jiang and Daumé III (2010)'s hybrid BP (`Jiang's method`), where we extract a solution by maximizing the max-marginals or sum-marginals of the max nodes. Note that these three message passing algorithms are "non-iterative"; since the hidden Markov chain is a tree, they terminate in a number of steps equal to the graph diameter. In some sense, this fact suggests that their power to solve the (still NP-hard) marginal MAP problem should be limited. We also implemented standard EM, starting from 10 random initializations and picking the best solution.

We show in Fig. 3(a) the percentage of correct solutions (configurations of all max nodes) for different algorithms, and Fig. 3(b) their relative energy errors as defined by $Q(\hat{x}_B; \theta) - Q(x_B^*; \theta)$, where $x_B$ is the estimated solution and $x_B^*$ is the true optimum. Our `Mix-Bethe` returns the highest percentage ($\geq 80\%$) of correct solutions across the range of $\sigma$, followed by $\text{Mix} - \text{TRW2}$. Surprisingly, `Mix-TRW1` usually works less well than `Mix-Bethe` and `Mix-TRW2` despite having better theoretical properties. This can be explained by the fact that `Mix-TRW1` rarely returns integer solutions. In general, we note that the truncated TRW approximations, including `Mix-TRW1` and `Mix-TRW2`, appear less accurate than the Bethe approximation (also a well known fact for max- and sum- inference) but are able to provide upper bounds, shown in Fig. 3(b).

Fig. 3(c) shows the algorithms' behavior over time for $\sigma = 0.8$, where each iteration is one CCCP step for `Mix-Bethe` and `Mix-TRW2` and an EM step for the EM algorithm. The EM update, while monotonic, is easily stuck at sub-optimal points; experimentally we observed that EM always terminated after only a few (2-3) iterations.

**Ising grid.** We also tested the algorithms on a marginal MAP problem constructed by a $10 \times 10$ Ising grid, in which the max and sum nodes are distributed in a chessboard pattern (Fig. 4). Note that in this case the sum graph (shaded) is not a tree.

We generate parameters randomly as before, but for binary states, giving "mixed" (attractive and repulsive) potentials; we also generated attractive-only potentials by taking the absolute value of $\theta_{ij}$. Since $G$ is no longer a tree, `Max-Product`, `Sum-product` and `Jiang's method` may fail to converge, in which case we tried adding damping (0.1) and additional iterations (200). We also ran a convergent alternative: CCCP for `Sum-product`, and sequential TRW for `Max-product`. Each algorithm reports its best result over all these options.

Fig. 4(b)-(c) shows the approximate relative error defined by $Q(x_B; \theta) - Q(\hat{x}_B; \theta)$, where $\hat{x}_B$ is best solution that we find across all algorithms. `Mix-Bethe` performs very well across all $\sigma$ for both mixed and attractive couplings. For attractive couplings, the three message passing algorithms also perform well; this is probably because the attractive models typically have two dominant modes, making the problem easier.

## 9 CONCLUSION

We have presented a general variational framework for solving marginal MAP problems approximately. Theoretically, our algorithms are justified by showing conditions under which the solutions are global or local optima. Our experiments demonstrate that our truncated Bethe approximation performs extremely well compared to similar approaches.

Future directions include improving the performance of the truncated TRW approximation by optimizing weights, deriving optimality conditions that may be applicable even when the sum component does not form a tree, studying mean field-like approximations, and extending these algorithms to "generalized" message passing on higher order cliques.

**Acknowledgments**

We thank the reviewers for their detailed feedback. This work was supported in part by NSF grant IIS-1065618.

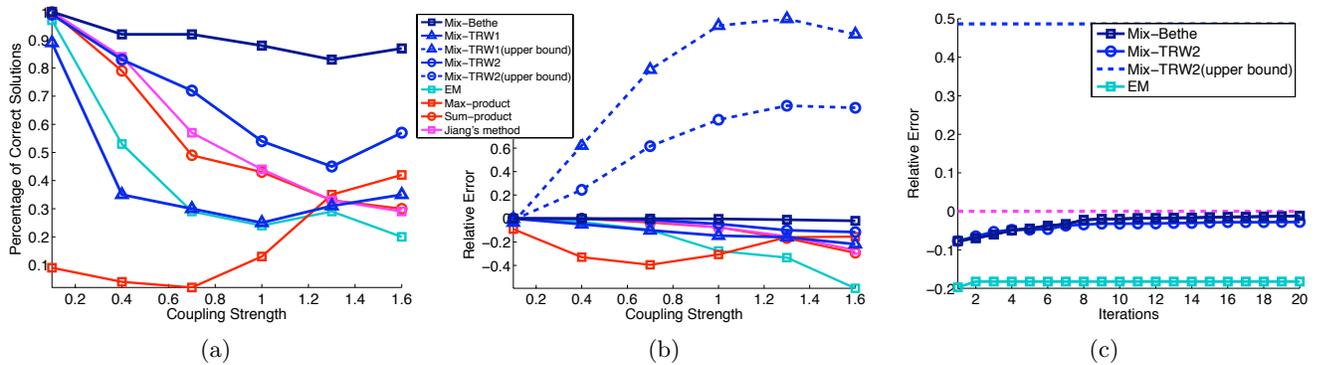

Figure 3: Results on the hidden Markov chain of Fig. 1 (best viewed in color). (a) Percentage of correct solutions for different algorithms. `Mix-Bethe` performs best across all coupling strengths. (b) The relative error of the solutions obtained by different algorithms, and the upper bounds obtained by `Mix-TRW1` and `Mix-TRW2`. (c) Performance as a function of iteration for $\sigma = .8$. EM is easily stuck in local optima and improves little over its initial guess.

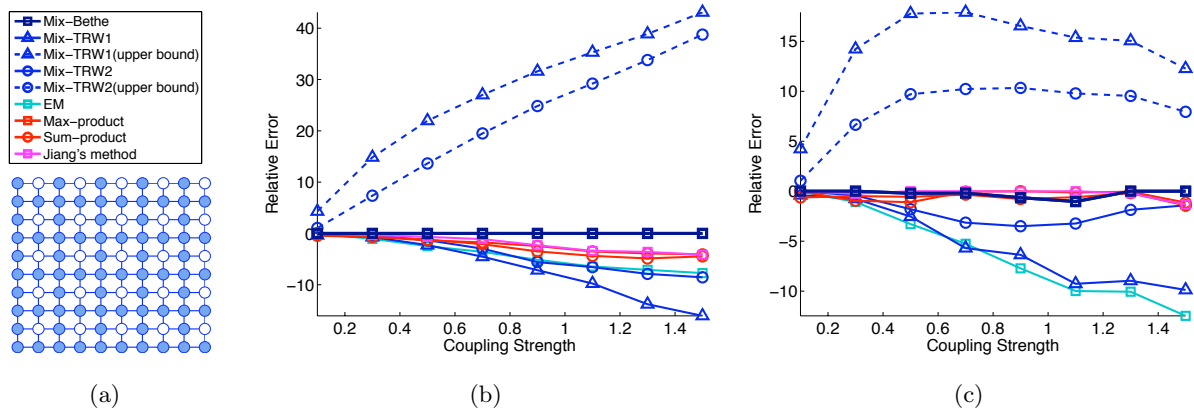

Figure 4: A marginal MAP problem defined on (a) a $10 \times 10$ Ising grid, with shaded sum nodes and unshaded max nodes; note the summation part is not a tree. (b) The approximate relative error of different algorithms for mixed potentials, as a function of coupling strength. `Mix-Bethe` always performs best in this case. (c) The approximate relative error for attractive potentials, as a function of coupling strength. In this case, the three message passing algorithms perform similarly to `Mix-Bethe`.